%% file: main.tex
\documentclass[review,journal]{IEEEtran}
\usepackage{amsmath,amssymb,amsfonts}
\usepackage{algorithmic}
\usepackage{algorithm}
\usepackage{array}
\usepackage[caption=false,font=normalsize,labelfont=sf,textfont=sf]{subfig}
\usepackage{textcomp}
\usepackage{stfloats}
\usepackage{url}
\usepackage{verbatim}
\usepackage{graphicx}
\usepackage{multicol}
\input{defs}

\usepackage{xcolor}

\IEEEpubid{0000--0000/00\$00.00~\copyright~2021 IEEE}

\begin{document}

\title{ACERAC: Efficient reinforcement learning in fine time discretization} 
%
\author{Jakub~\L{}yskawa, Pawe\l{}~Wawrzy\'nski
\\
Warsaw University of Technology, Institute of Computer Science, Warsaw, Poland} 

%
%

\maketitle
              
\begin{abstract}
One of the main goals of reinforcement learning (RL) is to provide a~way for physical machines to learn optimal behavior instead of being programmed. However, effective control of the machines usually requires fine time discretization. The most common RL methods apply independent random elements to each action, which is not suitable in that setting. It is not feasible because it causes the controlled system to jerk, and does not ensure sufficient exploration since a~single action is not long enough to create a~significant experience that could be translated into policy improvement. In our view these are the main obstacles that prevent application of RL in contemporary control systems. To address these pitfalls, in this paper we introduce an RL framework and adequate analytical tools for actions that may be stochastically dependent in subsequent time instances. We also introduce an RL algorithm that approximately optimizes a~policy that produces such actions. It applies experience replay to adjust likelihood of sequences of previous actions to optimize expected $n$-step returns the policy yields. The efficiency of this algorithm is verified against four other RL methods (CDAU, PPO, SAC, ACER) in four simulated learning control problems (Ant, HalfCheetah, Hopper, and Walker2D) in diverse time discretization. The algorithm introduced here outperforms the competitors in most cases considered.  
\end{abstract}
\begin{IEEEkeywords}
Reinforcement learning, Actor-Critic, Experience Replay, Fine Time Discretization.
\end{IEEEkeywords}

\section{Introduction} 

The subject of this paper is reinforcement learning (RL) \cite{2018sutton+1}. This field offers methods of learning to make sequential decisions in dynamic environments. One application of such methods is the literal implementation of ``machine learning'', i.e., enabling machines and software to learn optimal behavior instead of being programmed.

The usual goal of RL methods is to optimize a~policy that samples an action based on the~current state of a~learning agent. The only stochastic dependence between subsequent actions is through state transition: the action moves the agent to another state, which determines the distribution of another action. The main analytical tools in RL are based on this lack of other dependence between actions.  For example, for a~given policy, its value function expresses the expected sum of discounted rewards the agent may expect starting from a~given state. The sum of rewards does not depend on actions taken before the given state has been reached. Hence, only the given state and the policy matter. 

Lack of dependence between actions beyond state transition leads to the following difficulties. In the physical implementation of RL, e.g., in robotics, the lack of dependence usually means that white noise is added to control actions. However, this makes control discontinuous and subject to constant rapid changes. In addition, this is often impossible to implement since electric motors to execute these actions can not change their output too quickly. Even if such control is possible, it requires large amounts of energy, makes the controlled system shake, and exposes it to damages. 
\IEEEpubidadjcol

Control frequency for real-life control systems can be much higher than that of simulated environments for which RL methods are designed. The typical frequency of the control signal for environments commonly used as benchmarks for RL algorithms ranges from 20 to 60 Hz \cite{2016brockman+6}, while the control frequency considered for real-life robots is 10 times higher, from 200 to 500 Hz \cite{1987Khosla} and can be even higher, up to 1000 Hz \cite{2013Schrimpf}. Therefore, finer time discretization should be considered to make RL more suitable for automation and robotics.

The lack of dependence between actions beyond state transition may also reduce the efficiency of learning as follows. Each action is then an~independent random experiment that leads to policy improvement. However, due to the limited accuracy of (act\mbox{ion-)}value function approximation, the consequences of a~single action may be hard to recognize. The finer the time discretization, the more serious this problem becomes. The consequences of a~random experiment distributed over several time instants could be more tangible and thus easier to recognize. 

Additionally, fine time discretization makes policy evaluation more difficult, as it requires accounting for more distant rewards. Technically, the discount factor needs to be larger, which makes learning more difficult for most RL algorithms \cite{2015francois+1}. The above problems are serious enough to prevent RL from wide applicability of RL in real life control systems.

To avoid the above pitfalls, we introduce in this paper a~framework in which an~action is both a~function of state and a~stochastic process whose subsequent values are dependent. In particular, these subsequent values can be autocorrelated, which makes the resulting actions close to one another. A~part of action trajectory can create a~distributed-in-time random experiment that leads to policy improvement. An~RL algorithm is also introduced that optimizes a~policy based on the above principles. 

The contribution of this paper may be summarized by the following points: 
\begin{itemize} 
\item 
A framework is introduced here in which a~policy produces actions based on the states and values of a~stochastic process. This framework is suited for the application of RL to optimization of control in physical systems, e.g., in robots. 
\item 
An ACERAC algorithm, based on Actor-Critic structure and experience replay, is introduced that approximately optimizes a~policy in the aforementioned framework. 
\item 
An extensive study is described here with four benchmark learning control problems (Ant, Half-Cheetah, Hopper, and Walker2D) at diverse time discretization. The performance of the ACERAC algorithm is compared using these problems with state-of-the-art RL methods. 
\end{itemize} 

This paper extends \cite{2020szulc+2} in several directions. We introduce here the notion of adjusted noise which is the input to the noise-value function. Also, when manipulating the policy parameter, the value of the noise-value function at the end of the action sequence is taken into account. The experimental study of the resulting algorithm is almost entirely new. 

The rest of the paper is organized as follows. The problem considered here is formulated in Sec.~\ref{sec:problem}. Another section overviews related literature. Sec.~\ref{sec:policy} introduces a~policy that produces autocorrelated actions along with tools for its analysis. Sec.~\ref{sec:alg} introduces the ACERAC algorithm that approximately optimizes this policy. Sec.~\ref{sec:experiments} presents simulations that compare the algorithm presented with state-of-the-art reinforcement learning methods. The last section concludes the paper. 

\section{Problem Formulation} 
\label{sec:problem} 

We consider here the standard Markov Decision Process (MDP) model \cite{2018sutton+1} in which an agent operates in discrete time $t=1,2,\dots$. At time $t$ the agent finds itself in a~state, $\state_t\in\stateSpace$, takes an action, $\ctrl_t \in \ctrlSpace$, receives a~reward, $r_t \in \real$, and is transited to another state, $\state_{t+1} \sim P_\state(\cdot | \state_t, \ctrl_t)$, where $P_\state$ is a~fixed but unknown conditional probability. 

The goal of the agent is to learn to designate actions to be able to expect at each $t$ the highest discounted rewards in the future. To ensure exploration, there is usually a random component introduced into the action selection.

We mainly consider the application of the MDP model to control physical devices. Therefore, we assume that both $\stateSpace$ and $\ctrlSpace$ are spaces of vectors of real numbers \cite{2021Liu+4}. We also assume fine time discretization typical for such applications, which means that designating actions the agent should account for rewards that are quite distant in terms of discrete-time steps in the future. This translates into a~discount parameter close to $1$, e.g., $\gamma\in(0.995,1)$. We require the reasons for the instability of learning with such a~large~$\gamma$ \cite{2015francois+1} to be overcome. 

To ensure applicability to control of physical machines, we require that the actions should generally be close for subsequent~$t$, even if they are random. Also, the learning should be efficient in terms of the amount of experience needed to optimize the agent's behavior. 

\section{Related Work} 
\label{sec:related-work} 

A general way to make subsequent actions close is the autocorrelation of the randomness on which these actions are based. Efficiency in terms of experience needed can be provided by experience replay. We focus on these concepts in the literature review below. 

\subsection{Stochastic dependence between actions} 

An autocorrelated stationary stochastic process, now referred to as Ornstein-Uhlenbeck (OU) process, was analyzed in \cite{1930uhlenbeck+1}. This process is the only autocorrelated Gaussian stochastic process that has the Markov property \cite{1942doob}.

A~policy with autocorrelated actions was analyzed in \cite{2015wawrzynski}. This policy was optimized by a~standard RL algorithm that did not account for the dependence of actions. In \cite{2017vanhoof+2} a~policy was analyzed whose parameters were incremented by the~OU stochastic process. Essentially, this resulted in autocorrelated random components of actions. In~\cite{2019korenkevych+3} a~policy is analyzed that produced an~action that was the~sum of the~OU noise and a~deterministic function of the state. However, no learning algorithm was presented in the paper that accounted for the specific properties of this policy. 

\subsection{Reinforcement learning for fine time discretization}

In \cite{2019tallec+2} RL in arbitrarily fine time discretization is analyzed. It is proven that RL based on the action-value function can not be effective when time discretization becomes sufficiently fine and note the importance of the dependence of the action noise in the next timesteps. In the aforementioned work RL algorithm called Deep Advantage Updating (DAU) for discrete actions and its variant for continuous actions (CDAU) are introduced. These methods are based on estimating the advantage function and are presented as immune to time discretization. They are based on Deep Q-Network (DQN)~\cite{2013mnih+6} and Deep Deterministic Policy Gradient (DDPG)~\cite{2016lillicrap+7} algorithms, respectively, and use the OU process as autocorrelated noise.

Integral Reinforcement Learning (IRL) is an approach to learning control policies for continuous-time environments. IRL is based on the assumption that  the control problem can be divided into a hierarchy of control loops \cite{2009vrabie+1}. This assumption is usually not satisfied in challenging tasks and thus IRL is not applicable to tasks with any state transition dynamics, only those belonging to a~certain relatively narrow class \cite{2019Guo+2}.

\subsection{Reinforcement learning with experience replay} 

The Actor-Critic architecture for RL was first introduced in \cite{1983barto+2}. Approximators were applied to this structure for the first time in \cite{1998kimura+2}. Basic on-line RL algorithms use consecutive events of the agent-environment interaction to update the policy. To boost the efficiency of these algorithms, experience replay (ER) can be applied, i.e., storing the events in a~database, sampling, and using them for policy updates several times per each actual event \cite{1992mahadevan+1}. ER was combined with the Actor-Critic architecture for the first time in~\cite{2009wawrzynski}.

However, the application of experience replay to Actor-Critic encounters the following problem. The learning algorithm needs to estimate the quality of a~given policy based on the consequences of actions that were registered when a~different policy was in use. Importance sampling estimators are designed to do that, but they can have arbitrarily large variances. In~\cite{2009wawrzynski} the problem was addressed with truncating density ratios present in those estimators. In~\cite{2016wang+6} specific correction terms were introduced for that purpose. 

Another approach to the aforementioned problem is to prevent the algorithm from inducing a~policy that differs too much from the one tried. This idea was first applied in Conservative Policy Iteration~\cite{2002kakade+1}. It was further extended in Trust Region Policy Optimization~\cite{2015schulman+4}. This algorithm optimizes a~policy with the constraint that the Kullback-Leibler divergence between this policy and the one being tried should not exceed a~given threshold. The K-L divergence becomes an~additive penalty in Proximal Policy Optimization algorithms, namely PPO-Penalty and PPO-Clip~\cite{2017schulman+4}. 

A~way to avoid the problem of estimating the quality of a~given policy based on the one tried is to approximate the action-value function instead of estimating the value function. Algorithms based on this approach are DQN~\cite{2013mnih+6}, DDPG~\cite{2016lillicrap+7}, and Soft Actor-Critic (SAC)~\cite{2018haarnoja+3}. Although OU noise was added to the action in the original version of DDPG, this algorithm was not adapted to this fact in any specific way. SAC uses white noise in actions and it is considered one of the most efficient in this family of algorithms. 

\section{Policy with autocorrelated actions} 
\label{sec:policy} 

In this section, we introduce a~framework for reinforcement learning where subsequent actions are stochastically dependent beyond state transition. We also design tools for the analysis of such a~policy. 

Let an~action, $\ctrl_t$, be designated as 
\Beq \label{def:pi} 
    \ctrl_t = \pi(\state_t,\xi_t;\Aparam), 
\Eeq
where $\pi$ is a~deterministic transformation, $\state_t$ is a~current state, $\Aparam$ is a~vector of trained parameters, and $(\xi_t)_{t=1}^\infty$ is a~stochastic process with values in $\real^d$. We require this process to have the following properties: 
\begin{itemize} 
\item 
Stationarity: The marginal distribution of $\xi_t$ is the same for each $t$. 
\item 
Zero mean: $E\xi_t = 0$ for each $t$. 
\item 
Autocorrelation decreasing with growing lag: 
\Beq
    E\xi_t^T\xi_{t+k} > E\xi_t^T\xi_{t+k+1} > 0 
    \;\text{for}\; k\geq0. 
\Eeq
Essentially that means that values of the process are close to each other when they are in close time instants. 
\item 
Markov property: For any $t$ and $k,l\geq0$, the conditional distributions 
\Beq \label{xi:markov:prop} 
    (\xi_{t},\dots,\xi_{t+k} | \xi_{t-1}, \dots, \xi_{t-1-l}) 
    \;\; \text{and} \;\;
    (\xi_{t},\dots,\xi_{t+k} | \xi_{t-1})
\Eeq
are the same. In words, dependence of future values of the process, $\xi_{t+k},k\geq0$, on its past is entirely carried over by $\xi_{t-1}$.
\end{itemize} 

Consequently, if only $\pi$ \eqref{def:pi} is continuous for all its arguments, and subsequent states $\state_t$ are close to each other, then the corresponding actions are close too, even though they are random. Because they are close, they are feasible in physical systems. Because they are random, they create a~consistent distributed-in-time experiment that can give a~clue to policy improvement. 

Below we analyze an example of $(\xi_t)$ that meets the above requirements. 

\begin{figure} 
  \begin{center}
    \includegraphics[width=0.48\textwidth]{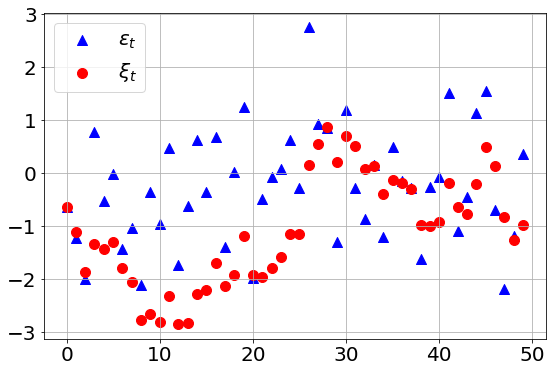}
  \end{center}
  \vspace{-1em}
  \caption{Realization of the~normal white noise $(\epsilon_t)$, and the Ornstein-Uhlenbeck process $(\xi_t)$  \eqref{AR:xi}. }
  \label{fig:xi} 
\end{figure} 

\paragraph{Ornstein-Uhlenbeck (OU) process $(\xi_t)$} Let $\alpha \in [0,1)$, $C$ be a positively definite matrix, and 
\Beq \label{AR:xi} 
    \begin{split} 
    \epsilon_t & \sim N(0,C), \quad t=1,2,\dots \\ 
    \xi_1 & = \epsilon_1 \\ 
    \xi_t & = \alpha \xi_{t-1} + \sqrt{1-\alpha^2} \epsilon_t, \quad t=2,3,\dots 
    \end{split} 
\Eeq
Fig.~\ref{fig:xi} demonstrates a~realization of both the white noise $(\epsilon_t)$ and $(\xi_t)$. Let us analyze if $(\xi_t)$ has the required properties. Their derivations can be found in Appendix \ref{app:AR:props}. 

Both $\epsilon_t$ and $\xi_t$ have the same marginal distribution $N(0,C)$. Therefore, $(\xi_t)$ is stationary and zero-mean. Applying induction to \eqref{AR:xi} one obtains
$$
    E\xi_{t}\xi_{t+k}^T = \alpha^{|k|} C 
    \; \text{ and } \;
    E\xi_{t}^T\xi_{t+k} = \alpha^{|k|} \text{tr}(C) 
$$ 
for any $t,k$. Therefore, $(\xi_t)$ is autocorrelated, and this autocorrelation decreases with growing lag. Consequently, the values of $\xi_t$ are closer to one another for subsequent $t$ than the values of $\epsilon_t$, namely 
\Beqso
    E\|\epsilon_{t} - \epsilon_{t-1}\|^2
    & = E(\epsilon_{t} - \epsilon_{t-1})^T(\epsilon_{t} - \epsilon_{t-1}) 
    = 2\text{tr}(C) \\ 
    E\|\xi_{t} - \xi_{t-1}\|^2 
    & = E\left((\alpha\!-\!1)\xi_{t-1} + \sqrt{1-\alpha^2}\epsilon_t\right)^T \\ 
    & \qquad\times\left((\alpha-1)\xi_{t-1} + \sqrt{1-\alpha^2}\epsilon_t\right) \\ 
    & = (\alpha-1)^2 \text{tr}(C) + (1-\alpha^2) \text{tr}(C) \\ 
    & = (1-\alpha)2\text{tr}(C). 
\Eeqso

The Markov property of $(\xi_t)$ directly results from how $\xi_t$~\eqref{AR:xi} is computed. 

In fact, marginal distributions of the process $(\xi_t)$, as well as its conditional distributions, are normal, and their parameters have compact forms. Let us denote 
\Beq
    \bar\xi^n_t = [\xi_t^T, \dots, \xi_{t+n-1}^T]^T. 
\Eeq
The distribution of $\bar\xi^n_t$ is normal 
\Beq \label{Omega_0} 
    N(0, \Omega^n_0), 
\Eeq
where $\Omega^n_0$ \eqref{Omega0^n} is a~matrix dependent on $n$, $\alpha$, and $C$. The conditional distribution $(\bar\xi^n_t|\xi_{t-1})$ is also normal, 
\Beq \label{B,Omega_1}
    N(B^n\xi_{t-1}, \Omega^n_1), 
\Eeq
where both $B^n$ \eqref{B^n} and $\Omega^n_1$ \eqref{Omega1^n} are matrices dependent on $n$, $\alpha$, and $C$. 

\paragraph{Noise-value function} In policy \eqref{def:pi} there is a~stochastic dependence between actions beyond the dependence resulting from the state transition. Therefore, the traditional understanding of policy as distribution of actions conditioned on state does not hold here. Each action depends on the current state, but also previous states and actions. Analytical usefulness of the traditional value function and action-value function is thus limited. 

Our objective now is to define an analytical tool in the form of a~function that satisfies the following: 
\begin{enumerate}
\item[R1.]
A hard requirement: The function designates an~expected value of future discounted rewards based on entities that this expected value is conditioned on. 
\item[R2.] 
An efficiency requirement: A~small change of policy corresponds to a~small change of this function. While this is not necessary, it facilitates concurrent learning of the policy and this function approximation. 
\end{enumerate}

In order to meet the above requirements we introduce an~{\it adjusted noise}, $(\uu_t)_{t=1}^\infty$, as follows. $\uu_t$ and $\xi_t$ belong to the same space $\real^d$. Let  
\Beq \label{def:f} 
    f(\cdot ; \Aparam, \state) 
\Eeq
be a bijective function in $\real^d$ parameterized by $\Aparam$ and state. We have 
\Beq \label{f,f^-1}
    \begin{split} 
    \xi_{t-1} = &  f(\uu_{t-1};\Aparam, \state_t) \\ 
    \uu_{t-1} = & f^{-1}(\xi_{t-1};\Aparam, \state_t).  
    \end{split} 
\Eeq
Formally, we can apply $f$ to convert $\xi_{t-1}$ to $\uu_{t-1}$ and back whenever necessary. 

As an~analytical tool satisfying the aforementioned hard requirement R1, we propose the {\it noise-value function} defined as 
\Beq \label{def:W} 
    W^\pi(\uu,\state) 
    = E_\pi\left(\sum_{i\geq0} \gamma^i r_{t+i} \Big| \xi_{t-1} = f(\uu;\Aparam,\state), \state_t = \state\right). 
\Eeq
The course of events starting in time $t$ depends on the current state $\state_t$ and the value $\uu_{t-1}$. Because of the Markov property of $(\xi_t)$ \eqref{xi:markov:prop} and the direct equivalence between $\xi_{t-1}$ and $\uu_{t-1}$, the pair $\langle\uu_{t-1},\state_t\rangle$ is a~proper condition for the expected value of future rewards. 

To satisfy the aforementioned efficiency requirement R2, we design the $f$ function~\eqref{def:f} based on~$\pi$. It should make the distribution of an initial part of the action trajectory $(\ctrl_t, \dots)$ similar for given $\langle \uu_{t-1}, \state_t\rangle$, regardless of the policy parameter~$\Aparam$. Therefore, when $\Aparam$ changes due to learning, the arguments of the $W^\pi$ function~\eqref{def:W} still define similar circumstances in which the rewards start being collected. This prevents large changes in the shape of~$W^\pi$. An example of an~appropriate $f$ function is provided below in~\eqref{f:example}. 

We can consider the pair $\langle\uu_{t-1},\state_t\rangle$ a~state of an extended MDP. Therefore, the noise-value function has all the properties of the ordinary value function. In particular, we consider $n$-step look-ahead equation in the form 
\Beqs \label{W=n-step} 
    & W^\pi(\uu_{t-1},\state_t) \\
    & =\! E_\pi\!\bigg(\sum_{i=0}^{n-1} \!\gamma^i r_{t+i} 
    \!+\! \gamma^n W^\pi(f(\xi_{t+n-1};\Aparam,\state_{t+n}),\state_{t+n})
    \Big| \uu_{t-1},\state_t\!\bigg). \notag
\Eeqs
It says that the noise-value function is the~expected sum of several first rewards, and that the rest of them are also designated by the noise-value function itself. 

The algorithm introduced below manipulates the policy $\pi$ \eqref{def:pi} to make $n$-step sequences of registered actions more or less likely in the future. Let us consider 
\Beqso
    \bar\state^n_t & = [\state_{t}^T,\dots,\state_{t+n-1}^T]^T, \\ 
    \bar\ctrl^n_i & = [\ctrl_{t}^T,\dots,\ctrl_{t+n-1}^T]^T, 
\Eeqso
and 
\Beq \label{gen:bar:pi} 
    \bar\pi(\bar\ctrl^n_t|\bar\state^n_t, \xi_{t-1};\Aparam)
\Eeq
being a probability density of the action sequence $\bar\ctrl^n_t$ conditioned on the sequence of visited states $\bar\state^n_t$, the preceding noise value $\xi_{t-1}$, and the policy parameter~$\Aparam$. This density is defined by $\pi$, and the conditional probability distribution $\bar\xi^n_t|\xi_{t-1}$. The algorithm defined in the next section updates $\Aparam$ to manipulate the above distribution.

\paragraph{The neural-AR policy} A simple and practical way to implement $\pi$ \eqref{def:pi} is as follows. A~feedforward neural network, 
\Beq
    A(\state;\Aparam), 
\Eeq
has input $\state$ and weights $\Aparam$. An~action is designated as 
\Beq \label{ctrl=A+xi} 
    \ctrl_t = \pi(\state_t,\xi_t;\Aparam) = A(\state_t;\Aparam) + \xi_t,  
\Eeq
for $\xi_t$ in the form~\eqref{AR:xi}. Let us analyze the distribution $\bar\pi$ \eqref{gen:bar:pi}. In this order the density of the normal distribution with mean $\mu$ and covariance matrix $\Omega$ will be denoted by 
\Beq
    \varphi(\cdot\,;\mu,\Omega). 
\Eeq
Let us also denote 
\Beq
    \bar A(\bar\state^n_i;\Aparam) = 
    [A(\state_{t};\Aparam)^T,\dots,A(\state_{t+n-1};\Aparam)^T]^T. 
\Eeq 
It can be seen that the distribution $(\bar\ctrl^n_t|\bar\state^n_t, \xi_{t-1})$ is normal, namely $N(\bar A(\bar\state^n_t;\Aparam) + B^n\xi_{t-1}, \Omega^n_1)$, (see \eqref{Omega_0} and \eqref{B,Omega_1}). Therefore, 
$$
    \bar\pi(\bar\ctrl^n_t|\bar\state^n_i,\xi_{t-1};\Aparam) 
    = \varphi(\bar\ctrl^n_t;\bar A(\bar\state^n_t;\Aparam)+B^n\xi_{t-1},\Omega^n_1). 
$$

What is of paramount importance is the log-density gradient $\nabla_\Aparam\ln\bar\pi$. For $\bar\pi$ defined as \eqref{ctrl=A+xi} it may be expressed as 
\Beqs \label{nabla:ln:bar:pi} 
    & \nabla_\Aparam \ln \bar\pi(\bar\ctrl^n_t|\bar\state^n_i,\xi_{t-1};\Aparam) \\
    & = \nabla_\Aparam \bar A(\bar\state^n_t;\Aparam)(\Omega^n_1)^{-1}(\bar\ctrl^n_t - B^n\xi_{t-1}-\bar A(\bar\state^n_t;\Aparam)). 
    \notag
\Eeqs

The $f$ function \eqref{def:f} may have the form 
\Beq \label{f:example} 
    \begin{split} 
    \uu_{t-1} & = f^{-1}(\xi_{t-1};\Aparam,\state_t) = A(\state_t;\Aparam) + \alpha \xi_{t-1} \\ 
    \xi_{t-1} & = f(\uu_{t-1};\Aparam,\state_t) = \alpha^{-1}(\uu_{t-1} - A(\state_t;\Aparam)). 
    \end{split} 
\Eeq
Then $\uu_{t-1}$ is the expected value of $\ctrl_t$ given $\Aparam$, $\state_t$, and $\xi_{t-1}$. Consequently, this definition of $f$ delimits differences between noise-value functions of different policies. This is because $W^\pi(\uu,\state)$ means for any policy the expected sum of future rewards received starting from the same point, which is the current state equal to $\state$ and the expected action equal to $\uu$. Therefore, if $W^\pi$ is accurately approximated for a~current policy and this policy is updated, the approximation of $W^\pi$ needs only limited adjustment. 

\section{ACERAC: Actor-Critic with Experience Replay and Autocorrelated aCtions} 
\label{sec:alg} 

The RL algorithm presented in this section has an actor-critic structure. It optimizes a~policy of the form~\eqref{def:pi} and uses the critic, 
$$
    W(\uu,\state;\Cparam), 
$$
which is an approximator of the noise-value function~\eqref{def:W} parametrized by the vector~$\Cparam$. The critic is trained to approximately satisfy \eqref{W=n-step}. 

A~constant parameter of the algorithm is natural~$n$. It denotes the length of action sequences whose probabilities the algorithm adjusts. For each time instant of the~agent-environment interaction, the policy~\eqref{def:pi} is applied. Also, data is registered that enables recall of the tuple $\langle \bar\state^n_t, \bar\ctrl^n_t, \bar\pi^n_t, \bar r^n_t, \state_{t+n} \rangle$, where $\bar\pi^n_t = \bar\pi(\bar\ctrl^n_t|\bar\state^n_t,\xi_{t-1};\Aparam)$. 

The general goal of policy training in ACERAC is to maximize $W^\pi(\uu_{j-1},\state_j)$ for each state $\state_j$ registered during the agent-environment interaction. In this order previous time instants are sampled, and sequences of actions that followed these instants are made more or less probable depending on their return. More specifically, $j$ is sampled from $\{t-M,\dots,t-n\}$, where $M$ is a~memory buffer length, and $\Aparam$ is adjusted along with a~policy gradient estimate, which is derived in Appendix~\ref{policy:gradient:est}. In other words, the conditional density of the sequence of actions $\bar\ctrl^n_j$ is being increased/dec\-reas\-ed depending on the return 
$$
    r_j + \dots + \gamma^{n-1} r_{j+n-1} + \gamma^{n} W(\uu_{j+n-1},\state_{j+n};\Cparam) 
$$
this sequence of actions yields. 

\subsection{Actor \& Critic training} 
At each $t$-th instant of agent--environment interaction experience replay is repeated several times in the form presented in Algorithm~\ref{alg:ACERAC} to calculate actor and critic weight updates. 

\begin{algorithm}
\caption{Calculating weights update from a single trajectory in Actor-Critic with Experience Replay and Autocorrelated aCtions, ACERAC}
\begin{algorithmic}[1]
\STATE {\textsc{ACTOR\_AND\_CRITIC\_UPDATES}$()$}
\STATE \hspace{0.5cm}$ \textbf{select randomly } j \in \{t-M \dots t-n\} $
\STATE \hspace{0.5cm}$ X_{j-1} \gets \mathbf{E}\left[\xi_{j-1} | \dots  \state_{j-1}, \ctrl_{j-1}, \dots;\Aparam\right]$
\STATE \hspace{0.5cm}$ X_{j+n-1} \gets \mathbf{E}\left[\xi_{j+n-1} | \dots \state_{j+n-1}, \ctrl_{j+n-1}, \dots ;\Aparam\right] $
\STATE \hspace{0.5cm}$ \uu_{j-1} \gets A(\state_j;\Aparam) + \alpha X_{j-1}$
\STATE \hspace{0.5cm}$ \uu_{j+n-1} \gets A(\state_{j+n};\Aparam) + \alpha X_{j+n-1}$
\STATE \hspace{0.5cm}$ d^n_j(\Aparam,\Cparam) = r_j + \dots + \gamma^{n-1} r_{j+n-1} $\par\hskip\algorithmicindent\hspace{0.5cm}$ + \gamma^{n}W(\uu_{j+n-1},\state_{j+n};\Cparam) -W(\uu_{j-1},\state_j;\Cparam)$
\STATE \hspace{0.5cm}$ \rho_j \gets \psi_b\left(\frac{
\bar\pi(\bar\ctrl^n_j| \bar\state^n_j, X_{j-1}; \Aparam)} {
\bar\pi^n_j} \right)$
\STATE \hspace{0.5cm}$ \Delta\Aparam \gets \nabla_\Aparam \ln \bar\pi(\bar\ctrl^n_j| \bar\state^n_j, X_{j-1}; \Aparam) d^n_j(\Aparam,\Cparam) \rho_j
$\par\hskip\algorithmicindent\hspace{0.5cm}$ + \gamma^n \nabla_\Aparam W(\uu_{j+n-1}(\Aparam),\state_{j+n};\Cparam) \rho_j
$\par\hskip\algorithmicindent\hspace{0.5cm}$
 - \nabla_\Aparam L(\state_j,\Aparam) $
\STATE \hspace{0.5cm}$\Delta\Cparam \gets \nabla_\Cparam W(\uu_{j-1}(\Aparam),\state_j;\Cparam) d^n_j(\Aparam,\Cparam) \rho_j(\Aparam) $
\STATE \hspace{0.5cm}\textbf{return }$\Delta\Aparam, \Delta\Cparam$
\end{algorithmic}
\label{alg:ACERAC}
\end{algorithm}

In Line 2, the algorithm selects an~experienced event to replay with the starting time index $j$. In the following lines the vectors of states $\bar\state^n_j = [\state_{j}^T,\dots,\state_{j+n-1}^T]^T$ and actions $\bar\ctrl^n_j = [\ctrl_{j}^T,\dots,\ctrl_{j+n-1}^T]^T$ are considered. 

In Lines 3-4 $X_{j-1}$ and $X_{j+n-1}$ are appointed to be values of the noise with which the current policy would designate the past actions. Then, in Lines 5-6, the corresponding adjusted noise values $\uu_{j-1}$ and $\uu_{j+n-1}$ values are calculated.

In Line 7 a temporal difference is computed. It determines the relative quality of~$\bar\ctrl^n_i$. 

In Line 8 a softly truncated density ratio is computed. The density ratio implements two ideas. Firstly, $\Aparam$ is changing due to being optimized, thus the conditional distribution $(\bar\ctrl^n_i|\xi_{j-1})$ is now different than it was at the time when the actions $\bar\ctrl^n_i$ were executed. The density ratio $\frac{
\bar\pi(\bar\ctrl^n_j| \bar\state^n_j, X_{j-1}; \Aparam)} {
\bar\pi^n_j}$ accounts for this discrepancy of distributions. Secondly, to limit the variance of the density ratio, the soft-truncating function~$\psi_b$ is applied. E.g., 
\Beq \label{soft:truncation} 
    \psi_b(x) = b\tanh(x/b), 
\Eeq
for a~certain~$b>1$. In the ACER algorithm~\cite{2009wawrzynski}, the hard truncation function, $\min\{\cdot,b\}$ is used for the same purpose which is limiting density ratios necessary in designating updates due to action distribution discrepancies. However, soft-truncating distinguishes the magnitude of density ratio and works slightly better than hard truncation. 

In Line 9 an improvement direction for actor is computed. The sum of $\nabla_\Aparam \ln \bar\pi(\bar\ctrl^n_j| \bar\state^n_j, X_{j-1}; \Aparam) d^n_j(\Aparam,\Cparam) \rho_j(\Aparam)$ and $\gamma^n \nabla_\Aparam W(\uu_{j+n-1}(\Aparam),\state_{j+n};\Cparam) \rho_j$ is an improvement direction estimate of $W^\pi(\uu_{j-1},\state_j)$ derived in Appendix \ref{policy:gradient:est}. It is designed to increase/decrease the likelihood of occurrence of the sequence of actions $\bar\ctrl^n_i$ proportionally to $d^n_i(\Aparam,\Cparam)$. $L(\state,\Aparam)$ is a~loss function that penalizes the actor for producing actions that do not satisfy constraints, e.g., they exceed their boundaries. 

In Line 10 an improvement direction for critic, $\Delta\Cparam$, is computed. It is designed to make $W(\cdot,\cdot\,;\Cparam)$ approximate the noise-value function~\eqref{def:W} better. 

In Line 7 the improvement directions $\Delta\Aparam$ and $\Delta\Cparam$ are applied to update $\Aparam$ and $\Cparam$, respectively, with the use of either ADAM, SGD, or another method of stochastic optimization. 

Implementation details of the algorithm using the neural-AR policy \eqref{ctrl=A+xi} are presented in \eqref{nabla:ln:bar:pi} and Appendix \ref{app:AR:props}. 

\section{Empirical study} 
\label{sec:experiments} 

This section presents simulations whose purpose was to compare the algorithm introduced in Sec.~\ref{sec:alg} to state-of-the-art reinforcement learning methods. 
We compared the new algorithm to Actor-Critic with Experience Replay (ACER) \cite{2009wawrzynski}, Proximal Policy Optimization (PPO) \cite{2017schulman+4}, Soft Actor-Critic (SAC) \cite{2018haarnoja+3} and Continuous Deep Advantage Updating (CDAU) \cite{2019tallec+2}.  We selected these algorithms as different state-of-the-art approaches that also apply trajectory updates (PPO, CDAU) or control exploration (SAC, CDAU). 
We used the RLlib implementation \cite{2018liang} of SAC and PPO, and implementation of CDAU published by its authors\footnote{https://github.com/ctallec/continuous-rl}.  Our experimental software is available online.\footnote{https://github.com/lychanl/acerac}

\begin{figure}
    \centering
    \subfloat[]{\includegraphics[width=0.47\linewidth]{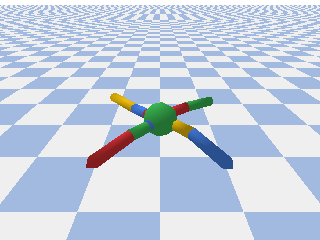}\label{fig:env:ant}}
    \hfil
    \subfloat[]{\includegraphics[width=0.47\linewidth]{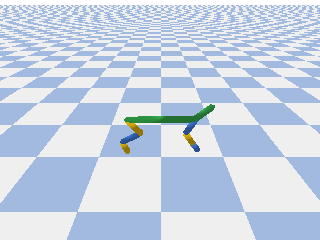}\label{fig:env:cheetah}} \\
    \subfloat[]{\includegraphics[width=0.47\linewidth]{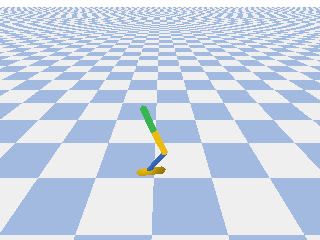}\label{fig:env:hopper}}
    \hfil
    \subfloat[]{\includegraphics[width=0.47\linewidth]{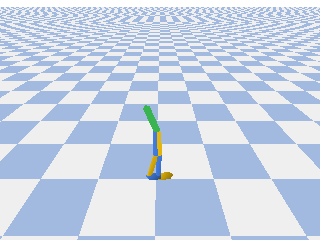}\label{fig:env:walker}}
    \caption{Environments used in simulations:  Ant (a), HalfCheetah (b), Hopper (c), Walker2D (d).}
    \label{fig:Envs}
\end{figure}

For the comparison of the RL algorithms to be the most informative we chose four challenging tasks inspired by robotics. They were Ant, Hopper, HalfCheetah, and Walker2D (see Fig.~\ref{fig:Envs}) from the PyBullet physics simulator \cite{2019coumans+1}. A~simulator that is more popular in the RL community is MuJoCo \cite{2012todorov}.\footnote{We chose PyBullet because it is freeware, while MuJoCo is commercial software.} Hyperparameters that assure optimal performance of ACER, SAC, and PPO applied to the environments considered in MuJoCo are well known. However, PyBullet environments introduce several changes to MuJoCo tasks, which make them more realistic and thus more difficult. Additionally, physics in MuJoCo and PyBullets differ slightly \cite{2015erez+2}, hence we needed to tune the hyperparameters. We based the hyperparameters in our experiments on their values used for MuJoCo environments reported in the original papers. We also followed their authors' guidelines when selecting hyperparameters for tuning. 

We do not limit the experiments only to the original environments. We also use modified ones with 3 and 10 times finer time discretization. This is to verify how the algorithms work in these circumstances. 

We used actor and critic structures as described in \cite{2018haarnoja+3} for each learning algorithm. That is, both structures had the form of neural networks with two hidden layers of 256 units each.

\subsection{Experimental setting} 

Each learning run with basic time discretization lasted for 3 million timesteps. Every 30000 timesteps of training a~simulation was made with frozen weights and without exploration for 5 test episodes. An average sum of rewards within a~test episode was registered. Each run was repeated 5 times. 

In experiments with, respectively, 3 and 10 times finer time discretization, the number of timesteps for a~run and between tests was increased, respectively, 3 and 10 times. Also, to keep the scale of the sum of discounted rewards, the discount parameter was increased from $0.99$ to, respectively, $0.99^{1/3}$ and $0.99^{1/10}$, and the rewards were decreased, respectively, 3 and 10 times. The number of model updates was kept constant for different discretization. The data buffer was increased 3 and 10 times, respectively. In ACER the $\lambda$ parameter was increased to $\lambda^{1/3}$ and $\lambda^{1/10}$, respectively. Also, in ACERAC, the $n$ coefficient was increased 3 and 10 times, respectively and $\alpha$ was increased to $\alpha^{1/3}$ and $\alpha^{1/10}$, respectively. 

For each environment-algorithm-discretization triple hyperparameters such as step-sizes were optimized to yield the highest ultimate average rewards. The values of these hyperparameters are reported in Appendix \ref{AlgorithmsHyperparams}. 

\begin{figure}
    \centering
    \includegraphics[width=0.9\linewidth]{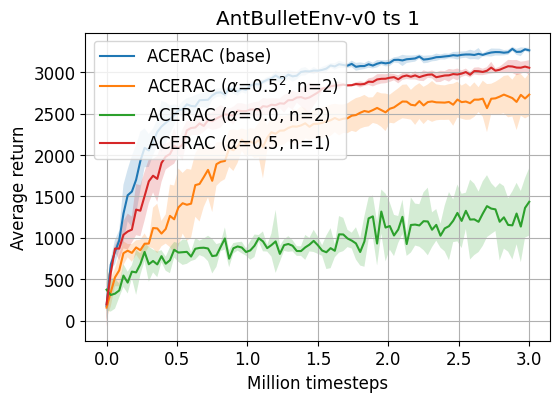}\\
    \includegraphics[width=0.9\linewidth]{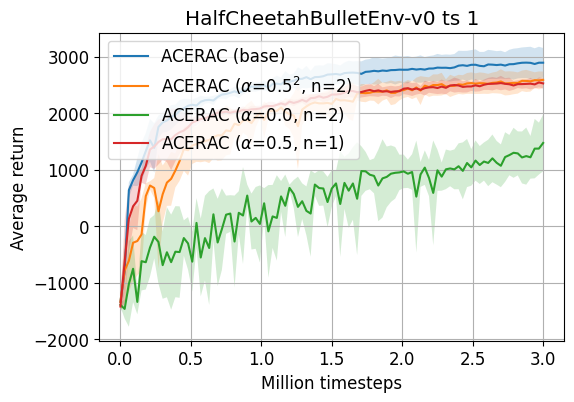}\\
    \includegraphics[width=0.9\linewidth]{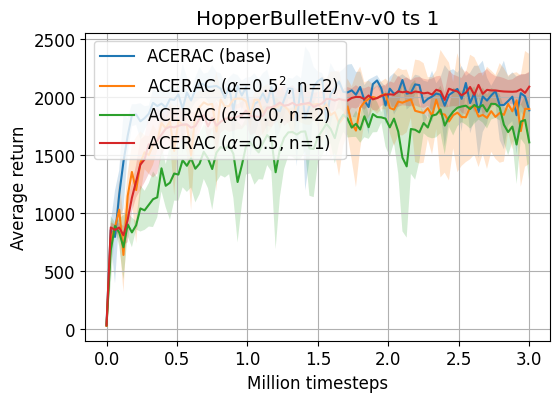}\\
    \includegraphics[width=0.9\linewidth]{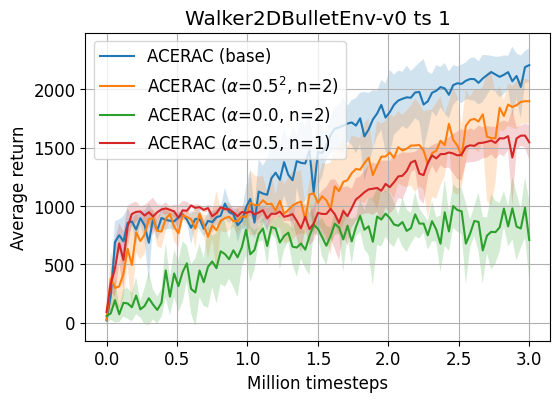}
    \caption{ACERAC with different $\alpha$ and $n$, for the original time discretization: Average sums of rewards in test trials. Base: $\alpha=0.5, n=2$. Environments: Ant, HalfCheetah, Hopper and Walker2D.  }
    \label{fig:AblationCurves_1}
\end{figure}

\begin{figure}
    \centering
    \includegraphics[width=0.9\linewidth]{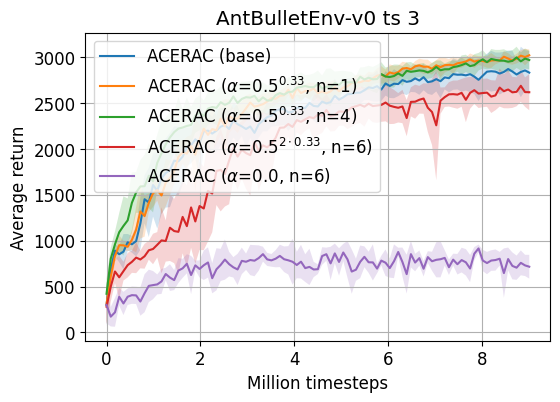}\\
    \includegraphics[width=0.9\linewidth]{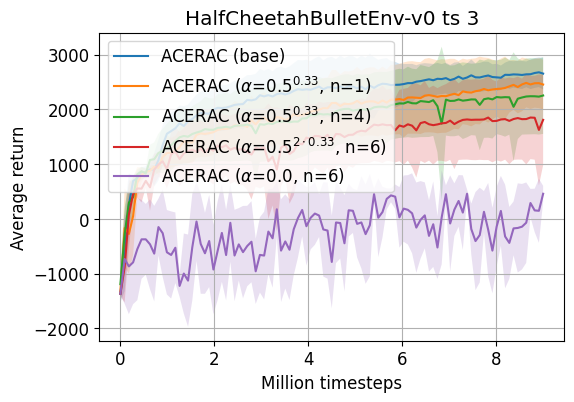}\\
    \includegraphics[width=0.9\linewidth]{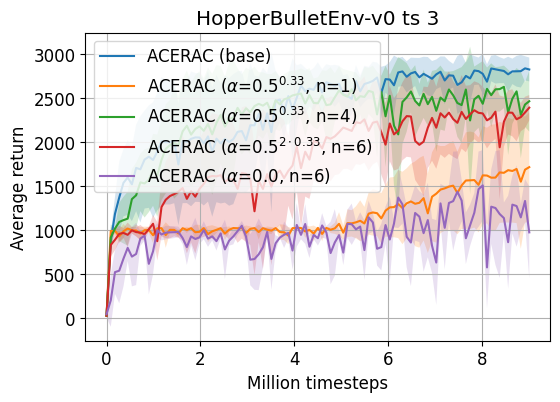}\\
    \includegraphics[width=0.9\linewidth]{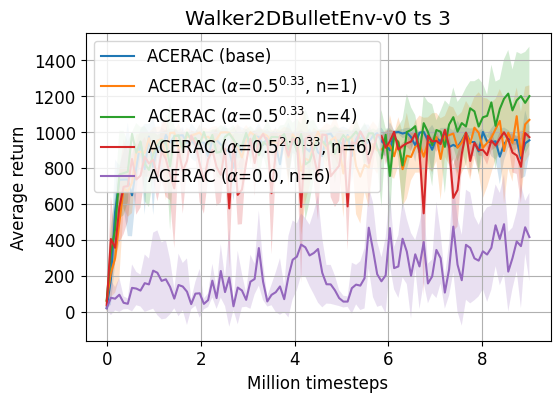}
    \caption{ACERAC with different $\alpha$ and $n$, for time discretization 3 times finer than the original: Average sums of rewards in test trials. Base: $\alpha=0.5^{1/3}, n=2\cdot3$. Environments: Ant, HalfCheetah, Hopper and Walker2D.  }
    \label{fig:AblationCurves_3}
\end{figure}

\begin{figure}
    \centering
    \includegraphics[width=0.9\linewidth]{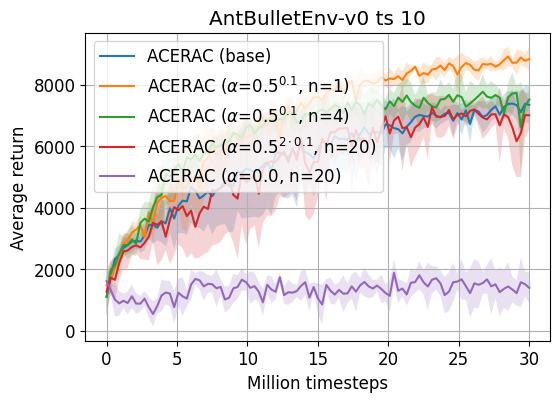}\\
    \includegraphics[width=0.9\linewidth]{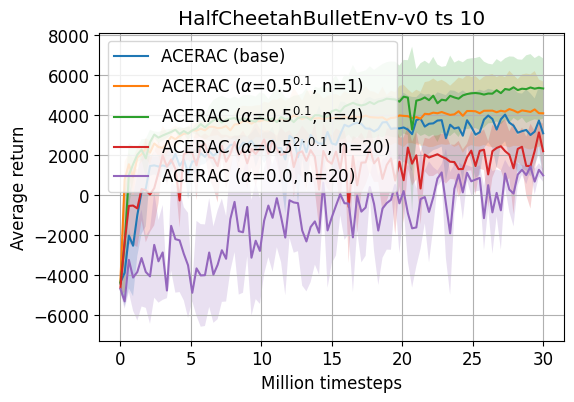}\\
    \includegraphics[width=0.9\linewidth]{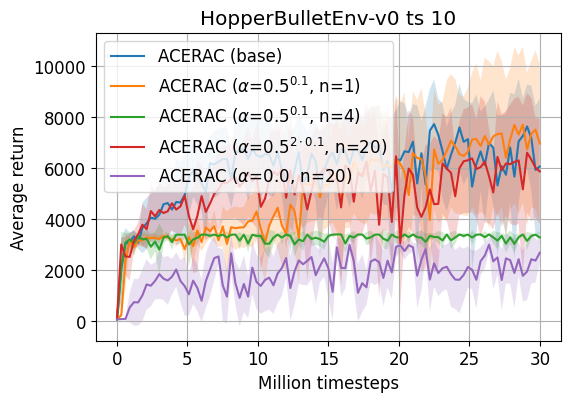}\\
    \includegraphics[width=0.9\linewidth]{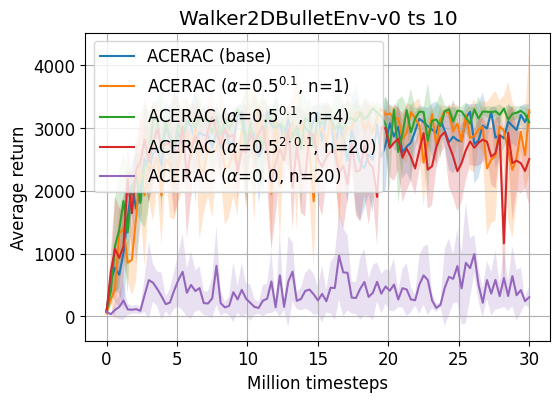}
    \caption{ACERAC with different $\alpha$ and $n$, for time discretization 10 times finer than the original: Average sums of rewards in test trials. Base: $\alpha=0.5^{1/10}, n=2\cdot10$. Environments: Ant, HalfCheetah, Hopper and Walker2D.  }
    \label{fig:AblationCurves_10}
\end{figure}

\subsection{Ablation results}

Figures \ref{fig:AblationCurves_1}, \ref{fig:AblationCurves_3}, and \ref{fig:AblationCurves_10}, respectively, present results for ACERAC, for the original, 3 times, and 10 times finer time discretization, with different $\alpha$ and $n$. The primary goal of these experiments was to verify whether the concepts introduced with ACERAC really contribute to performance. For $\alpha=0$, autocorrelation of actions is switched off. It is seen in the graphs that $\alpha=0$ yields inferior performance. $n$ defines the length of sequences of actions whose density is manipulated in the course of learning. It is seen that our proposed default values usually yield optimal or close to optimal performance, but in some cases smaller $n$ prove better.  

\begin{figure}
    \centering
    \includegraphics[width=0.9\linewidth]{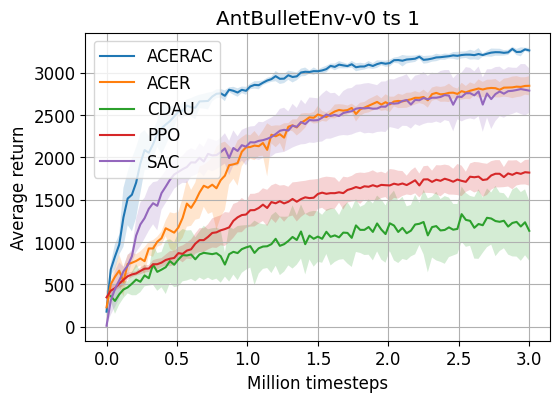}\\
    \includegraphics[width=0.9\linewidth]{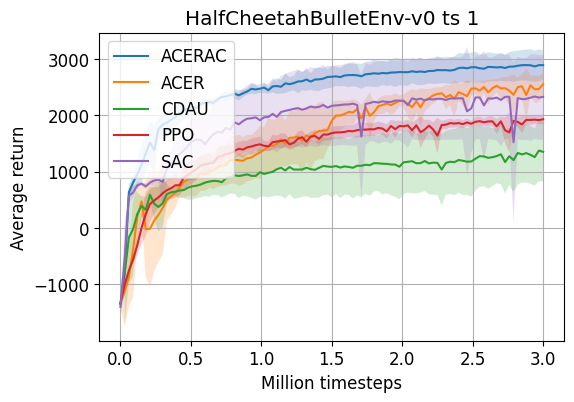}\\
    \includegraphics[width=0.9\linewidth]{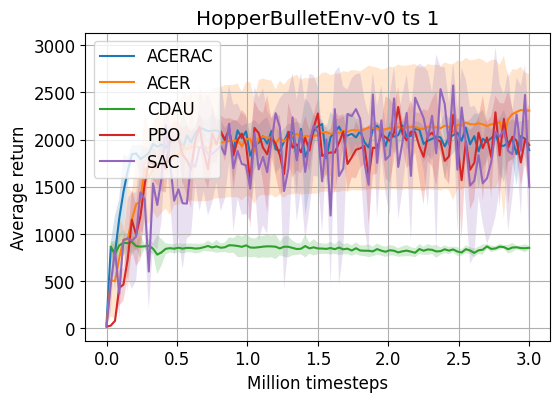}\\
    \includegraphics[width=0.9\linewidth]{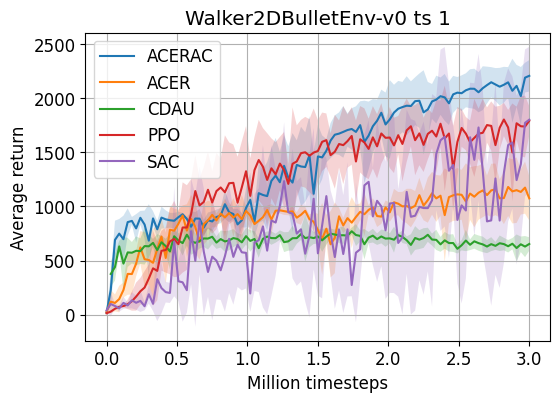}
    \caption{Learning curves for the original time discretization: Average sums of rewards in test trials. Environments: Ant, HalfCheetah, Hopper and Walker2D.  }
    \label{fig:Curves_1}
\end{figure}

\begin{figure}
    \centering
    \includegraphics[width=0.9\linewidth]{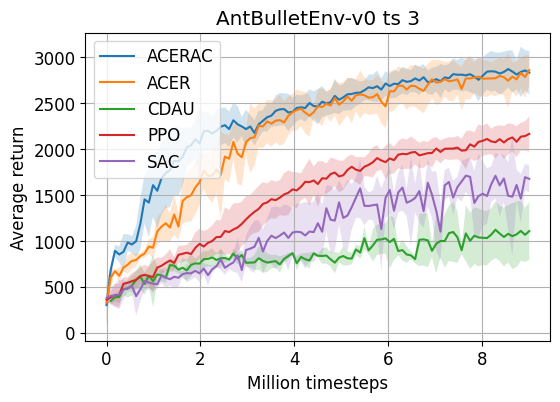}\\
    \includegraphics[width=0.9\linewidth]{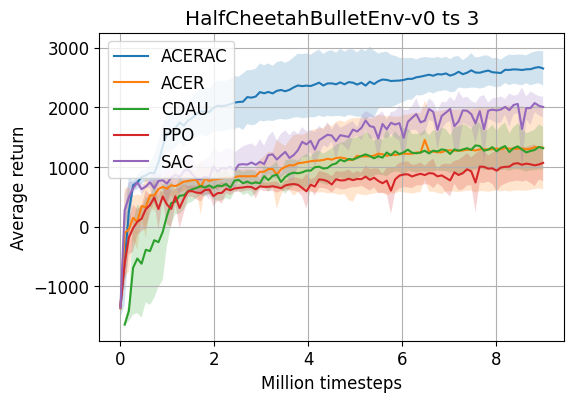}\\
    \includegraphics[width=0.9\linewidth]{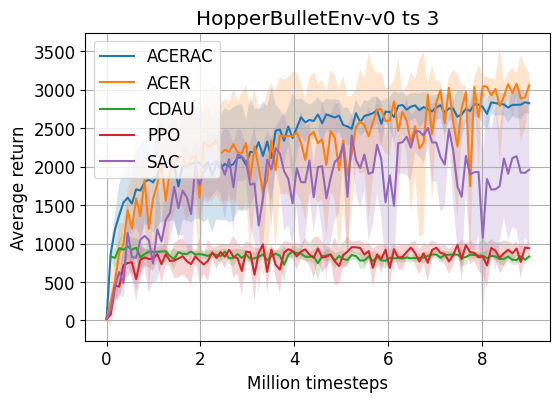}\\
    \includegraphics[width=0.9\linewidth]{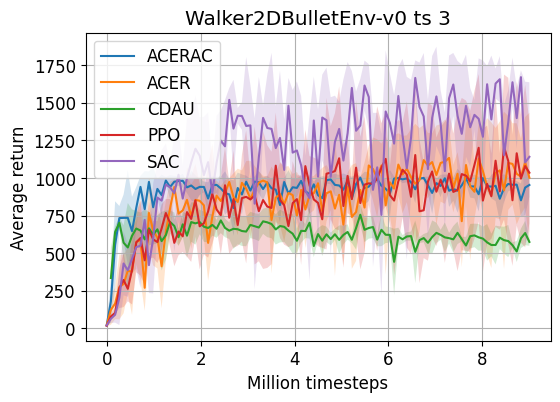}
    \caption{Learning curves for time discretization 3 times finer than the original: Average sums of rewards in test trials. Environments: Ant, HalfCheetah, Hopper and Walker2D.  }
    \label{fig:Curves_3}
\end{figure}

\begin{figure}
    \centering
    \includegraphics[width=0.9\linewidth]{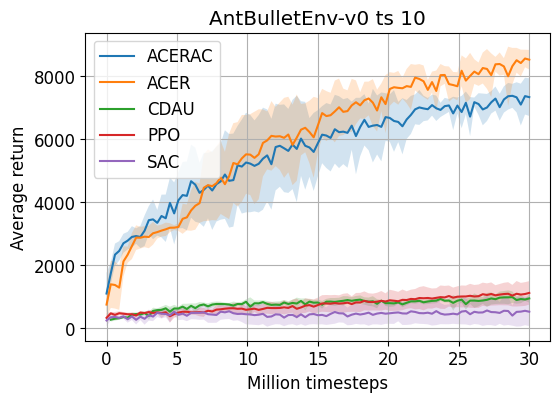}\\
    \includegraphics[width=0.9\linewidth]{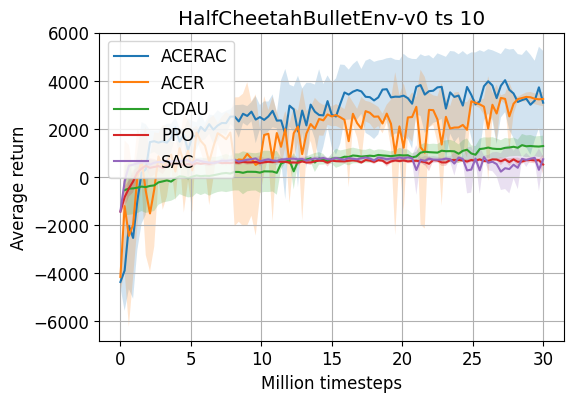}\\
    \includegraphics[width=0.9\linewidth]{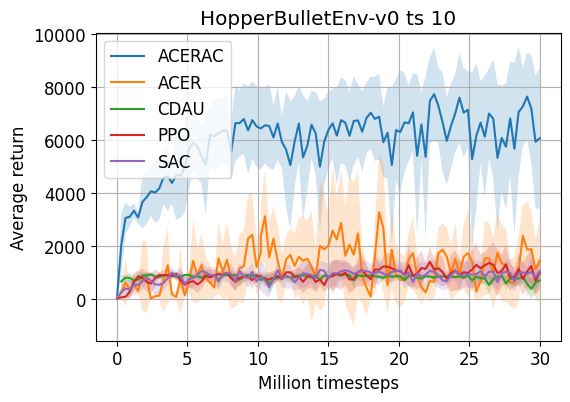}\\
    \includegraphics[width=0.9\linewidth]{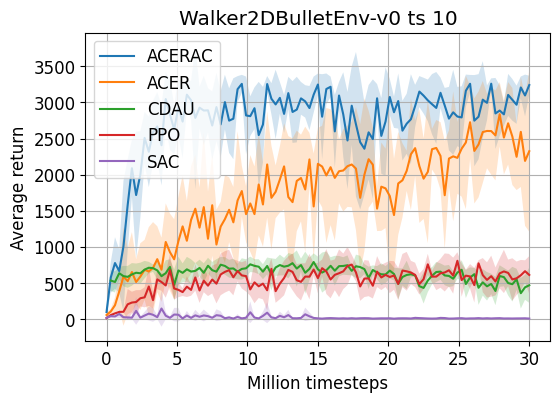}
    \caption{Learning curves for time discretization 10 times finer than the original: Average sums of rewards in test trials. Environments: Ant, HalfCheetah, Hopper and Walker2D.  }
    \label{fig:Curves_10}
\end{figure}

\subsection{Results} 

Figures~\ref{fig:Curves_1}, \ref{fig:Curves_3}, \ref{fig:Curves_10}, respectively, present learning curves for all four environments and all four compared algorithms. The figures are for, respectively, the original, 3 times, and 10 times finer time discretization. Each graph shows how a~sum of rewards, in test episodes evolves in the course of learning. Solid lines represent the average sums of rewards and shaded areas represent their standard deviations. 

It is seen in Figures~\ref{fig:Curves_1}--\ref{fig:Curves_10} that in 12 combinations of tasks and time discretizations, ACERAC was the algorithm to yield the best performance in 6 cases, in 2 cases it yielded the best performance {\it ex equo} with ACER, and in the rest of the cases it still yielded reasonable performance.

A curious result of our experiments was the extraordinarily high rewards obtained in some experiments with time discretization 10 times finer than the original. Namely, ACER and ACERAC obtained such results for Ant, and ACERAC for Hopper and Walker2D. Apparently, these environments require fast intervention of control and no algorithm is able to learn it at coarser time discretization.

It can also be seen that for most discretizations and problems ACERAC obtained relatively good results in the initial training steps, which is a desirable feature in robotic control \cite{1998Song+1}.

\subsection{Discussion} 

The performance of the algorithms in our experiments with fine time discretization can be attributed to two features. The first one is autocorrelated actions. ACERAC and CDAU use them, but only ACERAC utilize their properties. Other considered algorithms do not use them. It can be seen in Figures \ref{fig:Curves_1}-\ref{fig:Curves_10} that ACERAC achieved the best performance for 8 discretization-environment pairs out of 12. Switching off autocorelation from actions worsen the efficiency. The autocorrelated actions seem to be an~efficient way to organize exploration, better than actions without any stochastic dependence beyond state transition. However, a~policy with autocorrelated actions requires specialized training, which is provided in ACERAC.

The second factor is whether the algorithms use 1-step returns (SAC and CDAU) or $n$-step returns (PPO, ACER, and ACERAC). The impact of this parameter on performance is complex. For fine enough time discretization and large enough discount parameter, the 1-step returns are expected to fail due to the limited accuracy of the critic. However, if the critic is accurate enough for the task at hand, small $n$ values work quite well. This is visible in the upper part of Fig.~\ref{fig:AblationCurves_10}, where $1$ proves to be the best value of $n$ in ACERAC for Ant and the highest analyzed time discretization. Hence, $n>1$ may be a~remendy for an inaccurate critic, but an~accurate one is a~better remedy.  

Even though CDAU was designed in \cite{2019tallec+2} to assure efficient RL in fine time discretization, that algorithm yielded poor performance in our experiments. However, it was presented as an~extension of a~method, DAU, for discrete actions, and no experimental material on CDAU was presented in the original paper.  

\section{Conclusions and future work} 
\label{sec:conclusions} 

In this paper, a~framework has been introduced for the application of reinforcement learning to policies that admit stochastic dependence between subsequent actions beyond state transition. This dependence is a~tool that enables reinforcement learning in physical systems and fine time discretization. It can also yield better exploration and therefore faster learning. 

An algorithm based on this framework, Actor-Critic with Experience Replay and Autocorrelated aCtions (ACERAC), was introduced. Its efficiency was verified by simulations of four learning control problems, namely, Ant, HalfCheetah, Hopper, and Walker2D, at diverse time discretization. The algorithm was compared with CDAU, PPO, SAC, and ACER. ACERAC exhibited the best performance in 8 out of 12 discretization-environment pairs. 

It would be desirable to combine the framework proposed here with adapting the amount of randomness in actions by introducing reward for the entropy of their distribution, as is done in PPO. Also, the framework proposed here has been specially designed for applications in robotics. An obvious next step in our research would be to apply it in this area, which is more demanding than simulations.

\section*{Acknowledgement}

This work was partially funded by a grant of Warsaw University of Technology Scientific Discipline Council for Computer Science and Telecommunications.

This research was supported in part by PL-Grid Infrastructure.

\bibliographystyle{IEEEtran}
\input{references.bbl}


\appendix 

\input{appendix}

\end{document}

%% file: defs.tex
\def\state{s}
\def\stateSpace{\mathcal{S}}
\def\ctrl{a}
\def\ctrlSpace{\mathcal{A}}
\def\uu{u}

\def\Cparam{\nu}
\def\Aparam{\theta}
 
\def\real{\mathbb R}

\def\gd{\text d}
 
\def\comment#1{}

\def\eqref#1{(\ref{#1})}
\def\Beq#1\Eeq{\begin{equation}#1\end{equation}}
\def\Beqo#1\Eeqo{\begin{equation*}#1\end{equation*}}
\def\Beqs#1\Eeqs{\begin{align}#1\end{align}}
\def\Beqso#1\Eeqso{\begin{align*}#1\end{align*}}

%% file: references.bbl

%% file: appendix.tex
\subsection{Properties of Ornstein-Uhlenbeck process and the policy based on it} 
\label{app:AR:props} 

In this section the key properties of the process $(\xi_t)$ \eqref{AR:xi} are derived. 

\paragraph{\underline{Stationary distribution of $\xi_t$}}
From \eqref{AR:xi} one can see that if for a~certain~$t$ it is true that $\xi_{t-1}\sim N(0,C)$, then also $\xi_{t} \sim N(0,C)$. By induction this leads us to the conclusion that $\xi_t\sim N(0,C)$ for all~$t$. 

\paragraph{\underline{Stationary distribution of $\bar\xi^n_t$}}
Applying induction to \eqref{AR:xi} for $k\geq0$ one obtains that
\Beq \label{AR:xi_t+k}
    \xi_{t+k} = \alpha^k \xi_t + \sqrt{1-\alpha^2} \sum_{i=0}^{k-1} \alpha^i\epsilon_{t+k-i}. 
\Eeq
Consequently, 
$$
    E\xi_{t}\xi_{t+k}^T = \alpha^{|k|} C 
    \; \text{ and } \;
    E\xi_{t}^T\xi_{t+k} = \alpha^{|k|} \text{tr}(C) 
$$
Therefore, 
\Beqs
    \bar\xi_t^n & = [\xi_t^T,...,\xi_{t+n-1}^T]^T \sim N(0,\Omega_0^n) \notag \\ 
    \Omega_0^n & = \Lambda^n_0 \otimes C, \; 
    \Lambda^n_0 = [\alpha^{|l-k|}]_{l,k}, 0\leq l,k < n. \label{Omega0^n} 
\Eeqs
The symbol ``$\otimes$'' denotes Kronecker product of two matrices. We have 
\Beq
    (\Lambda^{n}_0 \otimes C)^{-1} = (\Lambda^{n}_0)^{-1} \otimes C^{-1}. 
\Eeq
\paragraph{\underline{Conditional distribution $\bar\xi_t^n|\xi_{t-1}$}} 
From \eqref{AR:xi_t+k} we have that 
$$
    E(\xi_{t+k}|\xi_{t-1}) = \alpha^{k+1} \xi_{t-1}, 
$$
and for $0\leq k\leq l$ we have 
\Beqso
    & \text{cov}(\xi_{t+k},\xi_{t+l} |\xi_{t-1}) \\ 
    & = E\left(\sqrt{1-\alpha^2} \sum_{i=0}^k\alpha^{k-i}\epsilon_{t+i}\right)
        \left(\sqrt{1-\alpha^2} \sum_{j=0}^l\alpha^{l-j}\epsilon_{t+j}\right)^T \\
    & = (1-\alpha^2) \alpha^{l-k} \left(1 + \alpha^2 + \dots + \alpha^{2k}\right) C \\ 
    & = \alpha^{l-k}\left(1-\alpha^{2k+2}\right) C. 
\Eeqso
Therefore, the conditional distribution $\bar\xi_t^n|\xi_{t-1}$ takes the form 
\Beqs
    \bar\xi_t^n | \xi_{t-1} & \sim N(B^n\xi_{t-1},\Omega_1^n) \label{xi^n|xi_t-1} \\ 
    B^n & = [\alpha I, \dots, \alpha^n I]^T \label{B^n} \\ 
    \Omega_1^n & = \Lambda^{n}_1\otimes C, \;
    \Lambda^{n}_1 = [\alpha^{|l-k|} - \alpha^{l+k+2}]_{l,k}, \;  0\leq l,k< n. \label{Omega1^n} 
\Eeqs

\paragraph{\underline{Distribution of actions' trajectory}}
For $\ctrl_t = A(\state_t;\Aparam) + \xi_t$ and $n>0$ we have $\bar\ctrl^n_t = \bar A(\bar\state^n_t;\Aparam) + \bar\xi^n_t$. The distribution of the actions that initiate a~trial, $\bar\pi(\bar\ctrl^n_t|\bar\state^n_j,\emptyset;\Aparam)$, is thus normal $N(\bar A(\bar\state^n_t;\Aparam),\Omega^n_0)$. The distribution of further actions $\bar\pi(\bar\ctrl^n_t|\bar\state^n_t,\xi_{t-1};\Aparam)$ is also normal, namely $N(\bar A(\bar\state^n_t;\Aparam)+B^n\xi_{t-1},\Omega^n_1)$. 
\paragraph{\underline{Retrieving $\xi_{t-1}$ and $\uu_{t-1}$ from past actions}} 
For the OU processes the values of $\xi_{t-1}$ and $u_{t-1}$ may be calculated from actions and actor's outputs as 
\Beqs
\xi_{t-1} & = \ctrl_{t-1} - A(\state_{t-1};\Aparam) \\ 
\uu_{t-1} &= A(\state_t;\Aparam) + \alpha\xi_{t-1} \\ 
& = A(\state_t;\Aparam) + \alpha(\ctrl_{t-1} - A(\state_{t-1};\Aparam)).
\Eeqs

If $t$ is an initial instance of a~trial, the conditional expected values of $\xi_{t-1}$ and $\uu_{t-1}$ are calculated from $A(\state_t;\Aparam)$ and $\ctrl_t$, namely
\Beqs
    \xi_{t-1} & = \alpha^{-1}(\ctrl_{t} - A(\state_t;\Aparam)) \\ 
    \uu_{t-1} & = A(\state_t;\Aparam) + \alpha\xi_{t-1} = \ctrl_t. 
\Eeqs

\subsection{Policy gradient estimator derivation} 
\label{policy:gradient:est} 

In this section we derive a policy gradient estimator, which is an estimator of a gradient of 
$$
    E_{\pi}\left(\sum_{i=0}^{n-1} \gamma^i r_{j+i} + \gamma^n W^\pi(\uu_{j+n-1}(\Aparam),\state_{j+n})\Big| \xi^*_{j-1}, \state_j \right)
$$
with respect to the current polity parameter $\Aparam$, for constant $\xi^*_{j-1} = \xi_{j-1}(\Aparam)$. 

Let us denote by $\ctrlSpace$ the action space, by $\Aparam$ the current policy parameter, by $\pi$ the current policy, by $\Aparam_j$ the policy parameter used when $\ctrl_j$ was selected, by $\pi(\Aparam_j)$ the policy used then, and the density ratio by 
$$
    \rho_j(\Aparam) = \frac{\bar\pi(\ctrl^n_j|\bar\state^n_j,\xi^*_{j-1};\Aparam)}{\bar\pi^n_j}.  
$$
We have 
\Beqso
  & \frac\gd{\gd\Aparam^T} E_{\pi(\Aparam)}\left(\sum_{i=0}^{n-1} \gamma^i r_{j+i} + \gamma^n W^\pi(\uu_{j+n-1}(\Aparam),\state_{j+n})\Big| \xi^*_{j-1}, \state_j \right) \\ 
  & = \frac\gd{\gd\Aparam^T} \int_{\ctrlSpace^n} \left(\sum_{i=0}^{n-1} \gamma^i r_{j+i} + \gamma^n W^\pi(\uu_{j+n-1}(\Aparam),\state_{j+n})\right) 
  \\ & \qquad \qquad \qquad \times \bar\pi(\bar\ctrl^n_j|\bar\state^n_j,\xi^*_{j-1};\Aparam) \gd\bar\ctrl^n_j \\ 
  & = \int_{\ctrlSpace^n} \left(\sum_{i=0}^{n-1} \gamma^i r_{j+i} + \gamma^n W^\pi(\uu_{j+n-1}(\Aparam),\state_{j+n}) \right) 
  \\ & \qquad\qquad \times 
  \nabla_\Aparam\bar\pi(\bar\ctrl^n_j|\bar\state^n_j,\xi^*_{j-1};\Aparam) \gd\bar\ctrl^n_j 
  \\ & \quad + \gamma^n \int_{\ctrlSpace^n} \nabla_\Aparam W^\pi(\uu_{j+n-1}(\Aparam),\state_{j+n}) 
  \bar\pi(\bar\ctrl^n_j|\bar\state^n_j,\xi^*_{j-1};\Aparam) \gd\bar\ctrl^n_j \\
  & = \int_{\ctrlSpace^n} \left(\sum_{i=0}^{n-1} \gamma^i r_{j+i} + \gamma^n W^\pi(\uu_{j+n-1}(\Aparam),\state_{j+n}) \right)
  \\ & \qquad\qquad \times 
  \frac{\nabla_\Aparam\bar\pi(\bar\ctrl^n_j|\bar\state^n_j,\xi^*_{j-1};\Aparam)}{\bar\pi(\bar\ctrl^n_j|\bar\state^n_j,\xi^*_{j-1};\Aparam)} \rho_j(\Aparam) \bar\pi^n_j \gd\bar\ctrl^n_j 
  \\ & \quad + \gamma^n \int_{\ctrlSpace^n} \nabla_\Aparam W^\pi(\uu_{j+n-1}(\Aparam),\state_{j+n}) 
  \rho_j(\Aparam) \bar\pi^n_j \gd\bar\ctrl^n_j \\
  & = E_{\pi(\Aparam_j)} \Bigg\{\Bigg[\left(\sum_{i=0}^{n-1} \gamma^i r_{j+i} + \gamma^n W^\pi(\uu_{j+n-1}(\Aparam),\state_{j+n}) \right)
  \\ & \qquad\qquad\qquad \times 
  \nabla_\Aparam\ln\bar\pi(\bar\ctrl^n_j|\bar\state^n_j,\xi^*_{j-1};\Aparam)
  \\ & \qquad\qquad\qquad  + \gamma^n \nabla_\Aparam W^\pi(\uu_{j+n-1}(\Aparam),\state_{j+n}) \Bigg] \rho_j(\Aparam) \Bigg\}. 
\Eeqso
The analytical property 
$$
    E_{\pi(\Aparam_j)} \left\{
    \nabla_\Aparam\ln\bar\pi(\bar\ctrl^n_j|\bar\state^n_j,\xi^*_{j-1};\Aparam)
    \rho_j(\Aparam) \right\} = 0
$$
allows us to subtract any constant {\it baseline} from the sum of rewards above. Consequently, an unbiased estimator of the policy gradient may take the form 
\Beqs 
& \Bigg[\!\!\left(\sum_{i=0}^{n-1}\!\gamma^i r_{j+i}\!+\! \gamma^n W^\pi(\uu_{j+n-1}(\Aparam),\state_{j+n}) \!-\! W^\pi(\xi^*_{j-1},\state_{j-1};\Cparam)\!\right) 
\notag \\ & \quad \times \label{unbiased:policy:grad:est} 
\nabla_\Aparam\ln\bar\pi(\bar\ctrl^n_j|\bar\state^n_j,\xi^*_{j-1};\Aparam) \\
  &\qquad + \gamma^n \nabla_\Aparam W^\pi(\uu_{j+n-1}(\Aparam),\state_{j+n}) \Bigg] 
  \rho_j(\Aparam). \notag  
\Eeqs
The above estimator is not feasible. Firstly, it is based on the noise-value function, which is unknown. Also, it uses the density ratio, which could make its variance excessive. In the feasible version of the above estimator, we use the approximator of the noise-value function, and the density ratio is softly truncated from above. 

\subsection{Algorithms' hyperparameters} 
\label{AlgorithmsHyperparams}

This section presents hyperparameters used in the simulations described in Sec.~\ref{sec:experiments}. For the original time discretization all algorithms used a~discount factor equal to $0.99$. Common parameters for the offline algorithms (i.e., ACERAC, ACER, SAC and CDAU) are presented in Tab.~\ref{tab:OfflineParams}. 
Hyperparameters specific for different algorithms are depicted in Tabs.~\ref{tab:ACERACParams}--\ref{tab:PPOSpecificParams}. The hyperparameters were tuned using grid search over values spread by a~factor of 3: $\dots, 10^{-6}, 3\cdot10^{-5}, 10^{-5}, \dots$, with the exception of the clip parameter for PPO, whose only considered values were $0.1$, $0.2$, $0.3$, as suggested by the authors of this algorithm \cite{2017schulman+4}.





\begin{table}
    \caption{Common parameters for offline algorithms (ACER, ACERAC, SAC, CDAU). $d$ denotes discretization increase (1, 3, or 10).}
    \centering
    \begin{tabular}{c|c}
        \hline
        Parameter & Value \\
        \hline
        Memory size & $d\cdot10^6$\\
        Minibatch size & 256 \\
        Update interval & $d$ \\
        Gradient steps & 1 \\
        \hline
    \end{tabular}
    \label{tab:OfflineParams}
\end{table}

\begin{table}
    \caption{ACERAC hyperparameters. $d$ denotes discretization increase (1, 3, or 10).}
    \centering
    \begin{tabular}{c|c}
        \hline
        Parameter & Value \\
        \hline
        Action std. dev. & 0.3 \\
        $\alpha$ & $0.5^{d^{-1}}$ \\ 
        Critic step-size & $10^{-4}$ \\
        Actor step-size & $10^{-5}$ \\
        $n$ & $d\cdot2$ \\
        $b$ & 2 \\
        Learning start & $d \cdot 10^3$ \\
        \hline
    \end{tabular}
    \label{tab:ACERACParams}
\end{table}

\begin{table}
    \centering
    \caption{ACER hyperparameters. $d$ denotes discretization increase (1, 3, or 10). For environment- and discretization-specific hyperparameters, see Tab.~\ref{tab:ACERLearningnRates}}
    \begin{tabular}{c|c}
        \hline
        Parameter & Value \\
        \hline
        Action std. dev. & 0.3 \\
        $\lambda$ & $1 - \frac{1 - 0.9}{d}$ \\
        $b$ & 2 \\
        Learning start & $d \cdot 10^3$ \\
        \hline
    \end{tabular}
    \label{tab:ACERParams}
    
\end{table}

\begin{table}
    \caption{SAC general hyperparameters. $d$ denotes discretization increase (1, 3, or 10). For environment- and discretization-specific hyperparameters see~Tab.~\ref{tab:SACScaling}}
    \centering
    \begin{tabular}{c|c}
        \hline
        Parameter & Value \\
        \hline
        Target smoothing coef. $\tau$ & 0.005 \\
        Learning start & $d \cdot 10^4$ \\
        \hline
    \end{tabular}
    \label{tab:SACParams}
\end{table}


\begin{table}
    \caption{PPO hyperparameters. $d$ denotes discretization increase (1, 3, or 10). For environment- and discretization-specific hyperparameters, see Tab.~\ref{tab:PPOSpecificParams}}
    \centering
    \begin{tabular}{c|c}
        \hline
        Parameter & Value \\
        \hline
        GAE parameter ($\lambda$) & 0.95 \\
        Minibatch size & 64 \\
        Horizon & $d \cdot 2048$ \\
        Number of epochs & 10 \\
        Value function clipping coef. & 10 \\
        Target KL & 0.01 \\
        \hline
    \end{tabular}
    \label{tab:PPOParams}
\end{table}

\begin{table}
    \caption{CDAU hyperparameters. $d$ denotes discretization increase (1, 3, or 10).}
    \centering
    \begin{tabular}{c|c}
        \hline
        Parameter & Value \\
        \hline
        dt & $d^{-1} \cdot 0.0165$ \\
        Step-size for HalfCheetah and Ant& $0.01$ \\ 
        Step-size for Hopper and Walker2D& $0.003$ \\ 
        $\theta$ & $7.5$ \\
        $\sigma$ & $1.5$ \\
        \hline
    \end{tabular}
    \label{tab:CDAUParams}
\end{table}

\begin{table}
    \caption{ACER step-sizes}
    \centering
    \begin{tabular}{c|c|c|c}
        \hline
        Parameter & \multicolumn{3}{c}{Value} \\
        \hline
        Discretization increase & 1 & 3 & 10 \\
        \hline
        Actor step-size for HalfCheetah env. & $10^{-5}$ & $3\cdot10^{-6}$ & $10^{-5}$ \\
        Actor step-size for Ant env. & $10^{-5}$ & $10^{-5}$ & $3\cdot10^{-6}$ \\
        Actor step-size for Hopper env. & $10^{-5}$ & $3\cdot10^{-5}$ & $3\cdot10^{-5}$ \\
        Actor step-size for Walker2D env. & $10^{-5}$ & $3\cdot10^{-5}$ & $10^{-6}$ \\
        Critic step-size for HalfCheetah env. & $10^{-5}$ & $3\cdot10^{-5}$ & $10^{-5}$ \\
        Critic step-size for Ant env. & $10^{-5}$ & $3\cdot10^{-5}$ & $10^{-4}$ \\
        Critic step-size for Hopper env. & $10^{-5}$ & $3\cdot10^{-5}$ & $10^{-5}$ \\
        Critic step-size for Walker2D env. & $10^{-5}$ & $3\cdot10^{-5}$ & $10^{-5}$ \\
    \end{tabular}
    \label{tab:ACERLearningnRates}
\end{table}
\begin{table}
    \caption{SAC reward scaling.}
    \centering
    \begin{tabular}{c|c|c|c}
        \hline
        Parameter & \multicolumn{3}{c}{Value} \\
        \hline
        Discretization increase & 1 & 3 & 10 \\
        \hline
        Reward scaling for HalfCheetah env. & 0.1 & 10 & 300 \\
        Reward scaling for Ant env. & 1 & 100 & 3000 \\
        Reward scaling for Hopper env. & 0.03 & 10 & 3 \\
        Reward scaling for Walker2D env. & 30 & 3 & 3 \\
        \hline
    \end{tabular}
    \label{tab:SACScaling}
\end{table}
\begin{table}
    \caption{PPO step-sizes and clip params.}
    \centering
    \begin{tabular}{c|c|c|c}
        \hline
        Parameter & \multicolumn{3}{c}{Value} \\
        \hline
        Discretization increase & 1 & 3 & 10 \\
        \hline
        Step-size for HalfCheetah env. & $3\cdot10^{-4}$ & $10^{-4}$ & $10^{-5}$ \\
        Step-size for Ant env. & $3\cdot10^{-4}$ & $3\cdot10^{-5}$ & $10^{-5}$ \\
        Step-size for Hopper env. & $3\cdot10^{-4}$ & $3\cdot10^{-4}$ & $10^{-5}$ \\
        Step-size for Walker2D env. & $3\cdot10^{-4}$ & $3\cdot10^{-4}$ & $10^{-5}$ \\
        Clip param for HalfCheetah env. & 0.2 & 0.3 & 0.1 \\
        Clip param for Ant env. & 0.2 & 0.2 & 0.1 \\
        Clip param for Hopper env. & 0.2 & 0.2 & 0.3 \\
        Clip param for Walker2D env. & 0.2 & 0.2 & 0.3 \\
        \hline
    \end{tabular}
    \label{tab:PPOSpecificParams}
\end{table}